\documentclass[letterpaper,journal]{IEEEtran}
\usepackage{amsmath,amsfonts}
\usepackage{algorithm}
\usepackage{array}
\usepackage[caption=false,font=normalsize,labelfont=sf,textfont=sf]{subfig}
\usepackage{textcomp}
\usepackage{stfloats}
\usepackage{url}
\usepackage{verbatim}
\usepackage{graphicx}
\usepackage{adjustbox}
\usepackage{amssymb}
\usepackage{times}
\usepackage{soul}
\usepackage{inputenc}
\usepackage{amsthm}
\usepackage{booktabs}
\usepackage[switch]{lineno}
\usepackage{color}
\usepackage{multirow}
\usepackage{float}
\usepackage{colortbl}
\usepackage[colorlinks,linkcolor=red,anchorcolor=black,citecolor=green]{hyperref}
\usepackage{algorithmicx}
\usepackage{algpseudocode}
\usepackage[table,xcdraw]{xcolor}

\usepackage[numbers,sort&compress]{natbib}
\usepackage{bbm}

\usepackage{caption}
\usepackage{subcaption} 

\begin{document}

\title{Toward Stable Semi-Supervised Remote Sensing Segmentation via Co-Guidance and Co-Fusion}

\author{
Yi Zhou, Xuechao Zou, Shun Zhang, Kai Li,~\IEEEmembership{Student Member, IEEE}, Shiying Wang, Jingming Chen, Congyan Lang, Tengfei Cao, Pin Tao, and Yuanchun Shi,~\IEEEmembership{Fellow, IEEE}

\thanks{ 
This work was partly supported by the Kunlun Talent and High-end Innovative and Entrepreneurial Talent program in Qinghai Province. Yi Zhou and Xuechao Zou have contributed equally to this work. Corresponding author: Shiying Wang.

Yi Zhou, Shiying Wang, Jingming Chen, and Tengfei Cao are with the School of Computer Technology and Application, Qinghai University, Xining, China. Shiying Wang and Tengfei Cao are also with the Intelligent Computing and Application Laboratory of Qinghai Province, Qinghai University, Xining, China (e-mail: yizhou.cs@foxmail.com; wangshiying.qhu@foxmail.com; chenjingming.qhu@foxmail.com; caotf@qhu.edu.cn).

Xuechao Zou, Shun Zhang, and Congyan Lang are with the Key Lab of Big Data \& Artificial Intelligence in Transportation (Ministry of Education), School of Computer Science \& Technology, Beijing Jiaotong University, Beijing, China (e-mail: xuechaozou@foxmail.com; xiaoshun3238@gmail.com; cylang@bjtu.edu.cn).

Kai Li, Pin Tao, and Yuanchun Shi are with the Department of Computer Science and Technology, Tsinghua University, Beijing, China. Pin Tao and Yuanchun Shi are also with the Key Laboratory of Pervasive Computing, Ministry of Education (e-mail: tsinghua.kaili@gmail.com; taopin@tsinghua.edu.cn; shiyc@tsinghua.edu.cn).}
}


\maketitle

\begin{abstract}
Semi-supervised remote sensing (RS) image semantic segmentation offers a promising solution to alleviate the burden of exhaustive annotation, yet it fundamentally struggles with pseudo-label drift, a phenomenon where confirmation bias leads to the accumulation of errors during training. In this work, we propose Co2S, a stable semi-supervised RS segmentation framework that synergistically fuses priors from vision-language models and self-supervised models. Specifically, we construct a heterogeneous dual-student architecture comprising two distinct ViT-based vision foundation models initialized with pretrained CLIP and DINOv3 to mitigate error accumulation and pseudo-label drift. To effectively incorporate these distinct priors, an explicit-implicit semantic co-guidance mechanism is introduced that utilizes text embeddings and learnable queries to provide explicit and implicit class-level guidance, respectively, thereby jointly enhancing semantic consistency. Furthermore, a global-local feature collaborative fusion strategy is developed to effectively fuse the global contextual information captured by CLIP with the local details produced by DINOv3, enabling the model to generate highly precise segmentation results. Extensive experiments on six popular datasets demonstrate the superiority of the proposed method, which consistently achieves leading performance across various partition protocols and diverse scenarios. Project page is available at \url{https://xavierjiezou.github.io/Co2S/}.
\end{abstract}


\begin{figure}[t]
    \centering
    \includegraphics[width=\columnwidth]{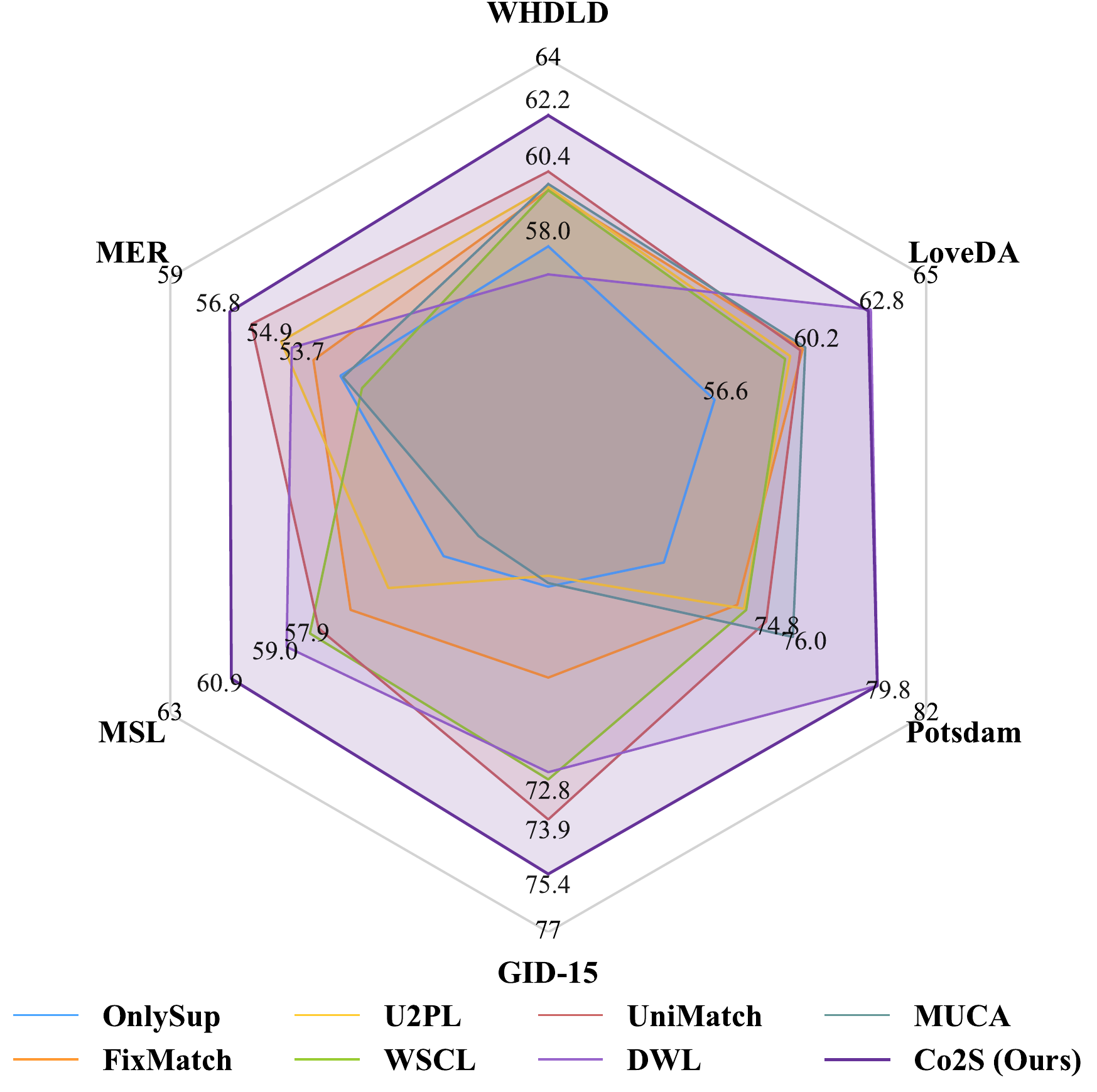}
    \caption{The radar chart compares the mIoU(\%) of different semi-supervised semantic segmentation methods across six remote sensing datasets under the 1/8 labeled ratio. Co2S consistently maintains leading performance across all benchmarks.}
    \label{fig:radar}
\end{figure}


\section{Introduction}\label{intro}

\IEEEPARstart{R}{emote} sensing imagery has become indispensable for a wide range of Earth observation applications, including land-cover mapping, urban planning, ecological monitoring, and disaster assessment~\cite{li2024review, huang2023deep, aleissaee2023transformers}. With the rapid advancement and widespread deployment of high-resolution satellite and aerial sensors, remote sensing data now exhibit large spatial coverage~\cite{xia2023openearthmap}, complex scene layouts~\cite{wang2022unetformer}, and fine-grained category structures~\cite{gao2021stransfuse}. Consequently, pixel-wise semantic segmentation has emerged as a core task, requiring models to be simultaneously sensitive to high-level semantics (e.g., category identity and contextual relationships) and low-level details (e.g., object boundaries and small structures~\cite{ma2021factseg}). Nevertheless, the fully supervised deep models that currently dominate this field heavily depend on exhaustive pixel-level annotations, which are notoriously expensive and time-consuming to obtain due to the need for expert interpretation~\cite{hua2021semantic}. In practice, only a small fraction of available images can be labeled, while the vast majority remain unlabeled~\cite{yu2023deep}. Under such conditions, there is a pressing need for semi-supervised segmentation methods that can effectively exploit abundant unlabeled data, cope with label scarcity, and maintain stable optimization despite the presence of noisy pseudo-labels~\cite{ran2024semi, zhu2025regionmatch}.

To mitigate the reliance on massive annotations, recent research has made substantial strides in semi-supervised learning, evolving through three primary paradigms in the remote sensing domain: generative adversarial networks (GANs)~\cite{hung2018adversarial,souly2017semi,mittal2019semi}, consistency-based methods~\cite{tarvainen2017mean,ke2019dual,chen2021semi,sohn2020fixmatch,yang2023revisiting}, and pseudo-labeling methods~\cite{yang2022st++,wang2022semi,jin2024dynamic,feng2022dmt}. Early GAN-based approaches, exemplified by Hung et al.~\cite{hung2018adversarial} and Souly et al.~\cite{souly2017semi}, align feature distributions via adversarial learning but often suffer from training instability and convergence difficulties. Subsequently, consistency-based methods like FixMatch~\cite{sohn2020fixmatch} and UniMatch~\cite{yang2023revisiting} became mainstream by enforcing prediction invariance under perturbations. However, they remain susceptible to confirmation bias, where incorrect predictions are reinforced over time. Similarly, pseudo-labeling strategies like U2PL~\cite{wang2022semi} and DWL~\cite{huang2024decouple} expand the training set with high-confidence predictions, yet they struggle with noise accumulation when the initial predictions are unreliable. Crucially, since these paradigms primarily rely on self-generated supervisory signals to guide training, they remain vulnerable to pseudo-label drift, particularly in complex remote sensing scenes. Without strong external guidance to rectify errors, these methods frequently encounter difficulties in distinguishing semantically similar categories or delineating precise object boundaries under severe label scarcity.

To tackle these challenges, we propose Co2S, a stable semi-supervised remote sensing segmentation framework. We construct a heterogeneous dual-student architecture that synergizes complementary priors derived from vision-language models and self-supervised models to mitigate pseudo-label drift. The framework is underpinned by an explicit-implicit semantic collaborative guidance mechanism and a global-local feature collaborative fusion strategy. Collectively, these components establish a stable mutual learning process, ensuring superior segmentation performance even under severe label scarcity.

The contributions of this work are summarized as follows:
\begin{enumerate}
     \item 
     We propose a stable dual-student learning framework for semi-supervised remote sensing segmentation, instantiated with two heterogeneous ViT-based vision foundation models. By leveraging distinct priors from pretrained CLIP and DINOv3, it enables reliable mutual learning and significantly alleviates pseudo-label drift.
    \item 
     We introduce an explicit-implicit semantic co-guidance mechanism that utilizes text embeddings and learnable queries to provide explicit and implicit class-level guidance, thereby jointly enhancing semantic consistency.
    \item
     We develop a global-local feature co-fusion strategy, effectively fusing the global contextual information captured by CLIP with the local texture details produced by DINOv3. This complementary fusion enables the model to generate highly precise segmentation results.
\end{enumerate}

Extensive experiments on six diverse remote sensing benchmarks, including WHDLD~\cite{shao2020multilabel}, LoveDA~\cite{wang2021loveda}, Potsdam~\cite{isprs-annals-I-3-293-2012}, GID-15~\cite{tong2020land}, MER~\cite{li2022stepwise}, and MSL~\cite{li2022stepwise}, demonstrate the effectiveness of Co2S.  The results show that our framework consistently delivers leading performance across various partition protocols, outperforming existing semi-supervised methods, particularly in regimes with extremely scarce annotations.

\section{Related Work}\label{sec: relatedwork}

\subsection{Remote Sensing Semi-Supervised Semantic Segmentation}

\noindent
In the early exploration phase, generative adversarial networks were incorporated into the Semi-Supervised segmentation framework to serve as an auxiliary supervision branch, facilitating model optimization through the adversarial discrimination of prediction distributions. Souly et al.~\cite{souly2017semi} leveraged a generator network to produce fake images using class-level information as extra training examples, forcing real samples to cluster in the feature space to improve pixel classification. Focusing on the output space, Hung et al.~\cite{hung2018adversarial} designed a discriminator in a fully convolutional manner to differentiate predicted probability maps from the ground truth, enabling the discovery of trustworthy regions in unlabeled data to facilitate effective semi-supervised learning.

Subsequently, consistency-based methods gained prominence by enforcing prediction invariance under perturbations. Meanwhile, FixMatch~\cite{sohn2020fixmatch} utilizes high-confidence predictions from weakly augmented images to supervise strongly augmented ones, providing a simplified yet effective means to leverage unlabeled data. Building on this, UniMatch~\cite{yang2023revisiting} introduces dual-stream input and feature perturbations to guide two strong views with a common weak view, successfully exploring a broader perturbation space for superior segmentation accuracy. Tailored for remote sensing, WSCL~\cite{lu2023weak} employs strong data augmentation and adaptive reweighting to decouple self-biased predictions and alleviate overfitting. Similarly, MUCA~\cite{wang2025semi} integrates multiscale uncertainty consistency and cross-teacher-student attention to construct discriminative feature representations, improving the distinction of highly similar objects. Parallel to consistency strategies, pseudo-labeling methods represent another active research direction, focusing on generating high-quality labels for self-training. U2PL~\cite{wang2022semi} enhances data utilization by identifying unreliable pixels based on entropy and using them as negative samples. Addressing the specific challenges of noise and class imbalance in remote sensing, DWL~\cite{huang2024decouple} employs a decoupled learning module alongside a ranking-based weighting module to adaptively filter and balance pseudo-labels. Different from these methods that rely on homogeneous co-training architectures or standard teacher-student paradigms, Co2S introduces a heterogeneous dual-student paradigm. By exploiting the complementary nature of distinct vision foundation models, our framework breaks the confirmation bias inherent in conventional co-training, offering an effective solution for pseudo-label drift.

\subsection{Semantic Segmentation with Vision Foundation Models}
\noindent
Vision foundation models have marked a paradigm shift in semantic segmentation, branching into vision-language models (VLMs) and self-supervised models (SSMs). VLMs~\cite{radford2021learning,rao2022denseclip,zhou2022extract,liu2024remoteclip,guo2024skysense} leverage text-image alignment for transferable visual understanding. CLIP~\cite{radford2021learning} utilizes large-scale contrastive pre-training to distill generic visual concepts, while methods like DenseCLIP~\cite{rao2022denseclip} and MaskCLIP~\cite{zhou2022extract} adapt it for dense prediction tasks. In remote sensing, RemoteCLIP~\cite{liu2024remoteclip} aligns visual features of remote sensing images with textual descriptions to learn robust visual features with rich semantics. Parallelly, SSMs~\cite{caron2021emerging,oquab2023dinov2,simeoni2025dinov3,sun2022ringmo,he2022masked,peng2022beit,zhou2021ibot,xie2022simmim} focus on extracting intrinsic spatial features, capturing hierarchical visual patterns without relying on manual annotations or textual supervision. The DINO family~\cite{caron2021emerging,oquab2023dinov2,simeoni2025dinov3} employs self-distillation to learn high-quality dense features effectively, whereas RingMo~\cite{sun2022ringmo} leverages generative self-supervised learning to specifically target dense and small objects in remote sensing scenes, Cloud-Adapter~\cite{cloud-adapter} proposes a parameter-efficient tuning strategy that adapts DINOv2~\cite{oquab2023dinov2} to the cloud segmentation without retraining the backbone.

Building on these advancements, recent studies explore architectures incorporating both VLM and SSM components~\cite{li2025exploring,zhang2025frozen,barsellotti2025talking}. Barsellotti et al.~\cite{barsellotti2025talking} proposed Talk2DINO, which aligns CLIP's textual embeddings with DINO's visual features via a learned mapping, resulting in less noisy segmentations with improved foreground distinction. For remote sensing, Ye et al.~\cite{ye2025towards} developed GSNet, employing query-guided feature fusion to integrate representations from generalist and specialist encoders to achieve more accurate mask predictions. Similarly, RSKT-Seg~\cite{li2025exploring} aggregates multi-directional cost maps to capture rotation-invariant cues, thereby enhancing both inference efficiency and segmentation accuracy. Alternatively, shifting the focus to utilizing frozen representations, Zhang et al.~\cite{zhang2025frozen} leveraged the static backbones of CLIP and DINO to generate high-quality pseudo-labels, enabling effective weakly supervised segmentation with reduced training costs. In a similar vein, CLIP-DINOiser~\cite{wysoczanska2024clip} applies DINO’s spatial affinities to refine the dense MaskCLIP features, producing smoother segmentation outputs without the need for pixel-level annotations. However, these approaches primarily target open-vocabulary supervised or weakly supervised scenarios. The potential of synergizing CLIP and DINO within a semi-supervised mutual learning framework, specifically to mitigate pseudo-label drift under label scarcity, remains unexplored. Our work bridges this gap by integrating their distinct priors to stabilize training when pixel-level annotations are severely limited.

\section{Method}

\begin{figure*}[t]
    \centering
    \includegraphics[width=\textwidth]{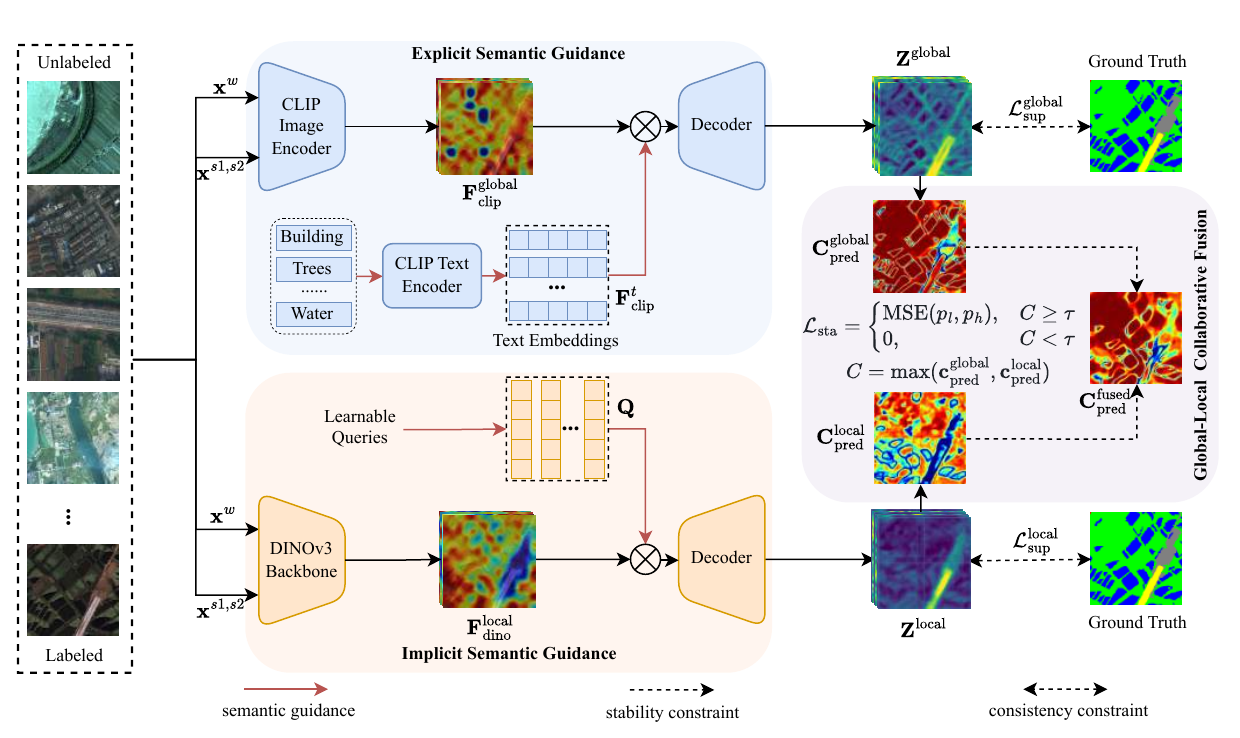}
    \caption{
        Overview of the proposed Co2S framework. 
        It integrates a CLIP-based student (top) using text embeddings for explicit semantic guidance and a DINOv3-based student (bottom) using learnable queries for implicit guidance. 
        For unlabeled data, the global-local collaborative fusion strategy enforces training stability by arbitrating supervision based on pixel-wise confidence.
    }
    \label{fig:Co2S_architecture} 
\end{figure*}

\subsection{Stable Dual-Student Semi-supervised Learning Framework}

\noindent
As illustrated in Fig.~\ref{fig:Co2S_architecture}, Co2S adopts a dual-student semi-supervised segmentation pipeline that leverages heterogeneous priors from CLIP~\cite{radford2021learning} and DINOv3~\cite{simeoni2025dinov3}. The two students process the same augmented views in parallel and output segmentation logits. Specifically, the CLIP student is guided by explicit class-level semantics encoded from text, while the DINOv3 student is driven by implicit class-level semantics represented by learnable queries. Both are injected into isomorphic yet parameter-independent query-based decoders (see Sec.~\ref{subsec:sem_guidance}). For labeled data, each student is supervised by the ground-truth mask. For unlabeled data, each student is trained with weak-to-strong consistency regularization across its own augmented predictions, while a global-local collaborative fusion mechanism is further introduced to perform confidence-driven pixel-wise mutual supervision between the two students, forming a stability constraint that suppresses pseudo-label drift (see Sec.~\ref{subsec:co_fusion}).

Formally, the training set consists of a labeled subset $\mathcal{D}_l=\{(\mathbf{x}_l,\mathbf{y}_l)\}$ and an unlabeled subset $\mathcal{D}_u=\{\mathbf{x}_u\}$, where $\mathbf{x}_l,\mathbf{x}_u\in\mathbb{R}^{3\times H\times W}$ and $\mathbf{y}_l\in\mathbb{R}^{1\times H\times W}$. Each student outputs segmentation logits with $N$ classes. For a labeled sample $(\mathbf{x}_l,\mathbf{y}_l)$, we generate a weakly augmented view $\mathbf{x}_l^w$ and feed it into both students, obtaining
$\mathbf{Z}^{\text{global}}\in\mathbb{R}^{N\times H\times W}$ from the CLIP student and
$\mathbf{Z}^{\text{local}}\in\mathbb{R}^{N\times H\times W}$ from the DINOv3 student. The supervised objective is applied independently:
\begin{equation}
\mathcal{L}_{\text{sup}}^{\text{global}}=\mathrm{CE}(\mathbf{Z}^{\text{global}},\mathbf{y}_l),\quad
\mathcal{L}_{\text{sup}}^{\text{local}}=\mathrm{CE}(\mathbf{Z}^{\text{local}},\mathbf{y}_l),
\end{equation}
where $\mathrm{CE}(\cdot,\cdot)$ denotes the pixel-wise cross-entropy between logits and the ground-truth mask, $\mathcal{L}_{\text{sup}}^{\text{global}}$ and $\mathcal{L}_{\text{sup}}^{\text{local}}$ denote the supervised losses for the CLIP and DINOv3 students, respectively.

For an unlabeled sample $\mathbf{x}_u$, we strictly adhere to the dual-stream perturbation paradigm of UniMatch~\cite{yang2023revisiting}. We generate a weakly augmented view $\mathbf{x}_u^w$, two strongly augmented views $\mathbf{x}_u^{s1}, \mathbf{x}_u^{s2}$, and one feature-perturbed view $\mathbf{x}_u^{fp}$, where $\mathbf{x}_u^w, \mathbf{x}_u^{s1}, \mathbf{x}_u^{s2}, \mathbf{x}_u^{fp} \in \mathbb{R}^{3 \times H \times W}$. For each student, let $\mathbf{Z}^w \in \mathbb{R}^{N \times H \times W}$ denote the segmentation logits predicted from the weak view, and $\mathbf{p}^w = \mathrm{Softmax}(\mathbf{Z}^w)$ denote the probability map, where $\mathrm{Softmax}(\cdot)$ is applied over the class dimension. For each pixel at location $(i, j)$, we assign the pseudo-label $\hat{y}_{u,i,j} = \arg\max_{n}(\mathbf{p}^w_{n,i,j})$ and the confidence score $c_{i,j} = \max_{n}(\mathbf{p}^w_{n,i,j})$, where $\hat{y}_{u,i,j} \in \{1, \dots, N\}$ and $c_{i,j} \in [0, 1]$. To formulate the consistency regularization, we define a generic masked cross-entropy function $\mathcal{H}$ given a logit map
$\mathbf{Z}\in \mathbb{R}^{N \times H \times W}$, a pseudo-label map
$\hat{\mathbf{y}}_u\in \mathbb{R}^{1 \times H \times W}$, and confidence scores
$\mathbf{c}\in \mathbb{R}^{1\times H\times W}$.
\begin{equation}
\mathcal{H}(\mathbf{Z}, \hat{\mathbf{y}}_u, \mathbf{c})
= \sum_{i,j} \mathbf{1}(c_{i,j} \ge \tau)\cdot \mathrm{CE}(\mathbf{Z}_{:,i,j}, \hat{y}_{u,i,j}),
\end{equation}
where $\mathbf{1}(\cdot)$ is the indicator function and $\tau$ is the confidence threshold. The final consistency loss $\mathcal{L}_{\text{ct}}$ is calculated as:
\begin{equation}
\mathcal{L}_{\text{ct}}
= \frac{1}{2} \mathcal{H}(\mathbf{Z}^{fp}, \hat{\mathbf{y}}_u, \mathbf{c})
+ \frac{1}{4}\Big( \mathcal{H}(\mathbf{Z}^{s1}, \hat{\mathbf{y}}_u, \mathbf{c})
+ \mathcal{H}(\mathbf{Z}^{s2}, \hat{\mathbf{y}}_u, \mathbf{c}) \Big),
\end{equation}
where $\mathbf{Z}^{fp}, \mathbf{Z}^{s1}, \mathbf{Z}^{s2}\in \mathbb{R}^{N \times H \times W}$ are the logits predicted from the
feature-perturbed view and the two strongly augmented views, respectively.

For each student, the total training objective is formulated as the weighted sum of the supervised, consistency, and stability loss:
\begin{equation}
\mathcal{L}
=
\lambda_{\text{sup}}\mathcal{L}_{\text{sup}}
+
\lambda_{\text{ct}}\mathcal{L}_{\text{ct}}
+
\lambda_{\text{sta}}\mathcal{L}_{\text{sta}}.
\end{equation}
Here, $\mathcal{L}_{\text{sup}}$ refers to $\mathcal{L}_{\text{sup}}^{\text{global}}$ for the CLIP student and $\mathcal{L}_{\text{sup}}^{\text{local}}$ for the DINOv3 student, $\mathcal{L}_{\text{ct}}$ and $\mathcal{L}_{\text{sta}}$ are defined analogously. We set $\lambda_{\text{sup}}=\lambda_{\text{ct}}=0.5$, and use a cosine ramp-up schedule for $\lambda_{\text{sta}}$. The detailed definition of stability loss $\mathcal{L}_{\text{sta}}$ is provided in Sec.~\ref{subsec:co_fusion}.

\subsection{Explicit-Implicit Semantic Collaborative Guidance}
\label{subsec:sem_guidance}

\noindent
To fully exploit heterogeneous pretraining, Co2S injects complementary semantic priors into the two students. The CLIP-based student receives explicit class-level guidance from a frozen text encoder, while the DINOv3-based student is driven by implicit class-level guidance via learnable queries.

For the CLIP-based student, we adopt a concept-based prompt ensembling strategy to improve semantic precision. For each semantic category $k\in\{1,\dots,N\}$, we curate a set of fine-grained concept descriptions
$\mathcal{C}_k=\{c_{k,i}\}_{i=1}^{S_k}$ (e.g., for \emph{road}: \{highway, main road, street, \dots\}), as specified in the dataset configurations, where $S_k$ denotes the number of concepts contained in $\mathcal{C}_k$. Each concept $c_{k,i} \in \mathcal{C}_k$ is formatted by a prompt template function $\mathcal{T}(\cdot)$ (e.g., $\mathcal{T}(c_{k,i})=\text{``a photo of a \{concept\}''}$) and then encoded by the frozen CLIP text encoder $\mathcal{E}_{\text{txt}}$. We aggregate these concept embeddings by averaging to obtain a class prototype $\mathbf{t}_k \in \mathbb{R}^{C_{1}}$:
\begin{equation}
\mathbf{t}_k
=
\frac{1}{S_k}\sum_{i=1}^{S_k}\mathcal{E}_{\text{txt}}\big(\mathcal{T}(c_{k,i})\big).
\label{eq:clip_concept_avg}
\end{equation}
Finally, we stack the prototypes of all $N$ categories into an explicit text-query matrix $\mathbf{F}_{\text{clip}}^{t} \in \mathbb{R}^{N \times C_{1}}$:
\begin{equation}
\mathbf{F}_{\text{clip}}^{t} = [\mathbf{t}_1, \dots, \mathbf{t}_N],
\label{eq:q_clip}
\end{equation}
which provides explicit class-level guidance for all categories using language-derived semantics.

In contrast, for the DINOv3-based student, we utilize learnable queries $\mathbf{Q} \in \mathbb{R}^{N \times C_{2}}$ to provide implicit class-level guidance~\cite{d2ls,rein}. These queries are implemented as a trainable embedding layer and are optimized alongside the DINOv3 backbone and the segmentation decoder:
\begin{equation}
\mathbf{Q}
=
[\mathbf{q}^{1},\dots,\mathbf{q}^{N}].
\label{eq:q_dino}
\end{equation}
Unlike the fixed text embeddings $\mathbf{F}_{\text{clip}}^{t}$, these learnable queries $\mathbf{Q}$ flexibly adapt to the underlying visual distribution of the imagery to encode implicit class-level embeddings.

Let $\mathbf{F}_{\text{clip}}^{\text{global}} \in \mathbb{R}^{C_{1} \times h \times w}$ and
$\mathbf{F}_{\text{dino}}^{\text{local}} \in \mathbb{R}^{C_{2} \times h \times w}$ denote the visual features from the
CLIP-based and DINOv3-based students, respectively. Both students employ an isomorphic yet parameter-independent query-based decoder, and compute
segmentation logits by dot-product similarity:
\begin{equation}
\mathbf{Z}^{\text{global}} = \mathrm{Sim}(\mathbf{F}_{\text{clip}}^{t}, \mathbf{F}_{\text{clip}}^{\text{global}}), \quad
\mathbf{Z}^{\text{local}}  = \mathrm{Sim}(\mathbf{Q}, \mathbf{F}_{\text{dino}}^{\text{local}}),
\label{eq:query_decoder}
\end{equation}
where $\mathbf{Z}^{\text{global}},\mathbf{Z}^{\text{local}}\in\mathbb{R}^{N\times h\times w}$, and $\mathrm{Sim}(\cdot,\cdot)$ denotes a dot-product similarity implemented with an \texttt{einsum} operator. This formulation maintains a unified architectural structure while allowing the two students to leverage distinct semantic sources, namely explicit text embeddings versus implicit learnable queries.

Although the CLIP text embeddings $\mathbf{F}_{\text{clip}}^{t}$ and the DINOv3 learnable queries $\mathbf{Q}$ are independent in the forward pass, they become coupled during optimization via the stability loss $\mathcal{L}_{\text{sta}}$. On mutually confident pixels, $\mathcal{L}_{\text{sta}}$ enforces consistency between the two predictions. Crucially, this mechanism enables the explicit class-level guidance from CLIP to rectify potential semantic category confusion in the DINOv3 student. Simultaneously, the implicit class-level guidance from DINOv3 reciprocally refines the coarse spatial localization of the CLIP student. Through this explicit and implicit semantic collaboration, Co2S effectively mitigates pseudo-label drift and stabilizes the dual-student learning process.

\begin{figure}[t]
    \centering
    \includegraphics[width=\columnwidth]{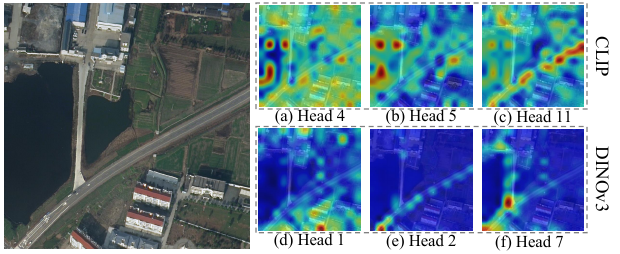}
    \caption{Visualization of attention maps from different heads of the CLIP image encoder (a-c) and DINOv3 backbone (d-f).}
    \label{fig:atten}
\end{figure}

\subsection{Global-Local Feature Collaborative Fusion}
\label{subsec:co_fusion}
\noindent
Conventional co-training frameworks~\cite{chen2021semi, qiao2018deep, ke2019dual} typically rely on identical network architectures with different random initializations. Despite the divergent starting points, these homogeneous models often converge to similar error patterns due to confirmation bias. To mitigate this error accumulation, Co2S introduces a global-local feature collaborative fusion strategy. As visually evidenced in Fig.~\ref{fig:atten}, the attention maps reveal distinct representational characteristics where CLIP produces diffuse activations rich in global contextual information while DINOv3 generates sharp responses highlighting local details. Building on this complementarity, our mechanism effectively fuses the global contextual information captured by CLIP~\cite{radford2021learning} with the local details produced by DINOv3~\cite{simeoni2025dinov3}. This complementary fusion enables the model to generate segmentation maps with more accurate semantics and sharper, well-defined boundaries.

Given an unlabeled image $\mathbf{x}_u$, its weakly augmented view is fed into the two students to obtain the global logits
$\mathbf{Z}^{\text{global}} \in \mathbb{R}^{N \times H \times W}$ and the local logits
$\mathbf{Z}^{\text{local}} \in \mathbb{R}^{N \times H \times W}$. We first compute the confidence maps
$\mathbf{C}_{\text{pred}}^{\text{global}} \in \mathbb{R}^{1\times H \times W}$ and
$\mathbf{C}_{\text{pred}}^{\text{local}} \in \mathbb{R}^{1 \times H \times W}$:
\begin{equation}
\mathbf{C}_{\text{pred}}^{\text{global}} = \max_{n}\big[\mathrm{Softmax}(\mathbf{Z}^{\text{global}})\big]_n,
\end{equation}
\begin{equation}
\mathbf{C}_{\text{pred}}^{\text{local}} = \max_{n}\big[\mathrm{Softmax}(\mathbf{Z}^{\text{local}})\big]_n,
\end{equation}
where $\mathrm{Softmax}(\cdot)$ is applied over the class dimension. Let
$c_{\text{pred}}^{\text{global}}$ and $c_{\text{pred}}^{\text{local}}$ denote the scalar confidence values at a specific spatial location
from $\mathbf{C}_{\text{pred}}^{\text{global}}$ and $\mathbf{C}_{\text{pred}}^{\text{local}}$, respectively. The joint maximum confidence for this pixel is defined as:
\begin{equation}
C = \max\big(c_{\text{pred}}^{\text{global}},\, c_{\text{pred}}^{\text{local}}\big).
\label{eq:conf_max}
\end{equation}

We then build interaction masks based on three confidence scenarios:
(1) If $c_{\text{pred}}^{\text{global}}\ge\tau$ and $c_{\text{pred}}^{\text{local}}\ge\tau$, the lower-confidence student learns from the higher-confidence one, when their confidence is equal, they engage in symmetric mutual learning.
(2) If $C \ge \tau$ but one of $\{c_{\text{pred}}^{\text{global}},\,c_{\text{pred}}^{\text{local}}\}$ is below $\tau$, the low-confidence student is supervised by the high-confidence peer.
(3) If $C < \tau$, the pixel is ignored and does not contribute to stabilization. Formally, the stability loss is defined as:
\begin{equation}
\mathcal{L}_{\text{sta}}(t)
= \lambda(t)\,
\begin{cases}
\text{MSE}\big(\mathbf{p}_{\ell},\, \mathbf{p}_{h}\big), & C \ge \tau,\\[3pt]
0, & C < \tau,
\end{cases}
\label{eq:l_sta}
\end{equation}
where $\text{MSE}(\cdot,\cdot)$ represents the mean squared error calculation, $\mathbf{p}_{\ell}, \mathbf{p}_{h} \in \mathbb{R}^{N}$ denote the class-probability vectors of the lower- and higher-confidence students at each pixel, respectively. They are computed by applying $\mathrm{Softmax}(\cdot)$ to the corresponding logits, $\tau$ is the confidence threshold, and $\lambda(t)$ is a ramp-up weight depending on the training step $t$. 

This pixel-wise fusion strategy bridges global contextual clues with local texture details based on pixel-wise confidence. By synergizing these distinct priors at the optimization level, the proposed strategy effectively compensates for limited supervision, yielding predictions that are both semantically stable and spatially precise even under scarce labels.

\section{Experiments}

\subsection{Datasets}

\noindent
We evaluate Co2S on six widely used remote sensing segmentation benchmarks, covering a diverse range of sensors, spatial resolutions, geographic regions, and scene complexities.

\subsubsection{WHDLD}
WHDLD~\cite{shao2020multilabel} comprises 4,940 optical images captured by GaoFen-1 and ZiYuan-3 satellites, each having a size of 256$\times$256 pixels and a 2-meter spatial resolution. It provides pixel-wise annotations for six categories: building, road, pavement, vegetation, bare soil, and water.

\subsubsection{LoveDA}
LoveDA~\cite{wang2021loveda} provides 5,987 images of size 1,024$\times$1,024 pixels with 0.3-meter resolution. Spanning urban and rural scenes, it annotates seven categories: background, building, road, water, barren, forest, and agriculture. 

\subsubsection{Potsdam}
Potsdam~\cite{isprs-annals-I-3-293-2012} consists of 38 aerial orthophotos with 5-cm resolution and 6,000$\times$6,000 dimensions. It covers six classes: impervious surface, building, low vegetation, tree, car, and clutter. These images are cropped into 5,472 non-overlapping 512$\times$512 patches for training and evaluation.
 
\subsubsection{GID-15}
GID-15~\cite{tong2020land} originates from ten GaoFen-2 satellite images with approximate dimensions of 7,200$\times$6,800 pixels. It provides pixel-level annotations for 15 fine-grained categories, spanning industrial, residential, agricultural, and water regions. These large images are tiled into 3,420 non-overlapping 512$\times$512 patches to facilitate large-scale land-cover classification.

\subsubsection{MER and MSL}
Derived from HiRISE imagery for Mars terrain analysis~\cite{li2022stepwise}, MER contains 1,023 grayscale images sized  1,024$\times$1,024, while MSL comprises 4,155 RGB images measuring 560$\times$500. They share nine distinct planetary surface categories: martian soil, sands, gravel, bedrock, rocks, tracks, shadows, unknown, and background.

\subsection{Implementation Details}

\noindent{The proposed Co2S framework is optimized using the AdamW optimizer~\cite{loshchilov2017decoupled} with a weight decay of $0.01$ and a polynomial learning rate decay schedule (power $= 0.9$). Structurally, both students utilize a ViT-B/16 backbone~\cite{dosovitskiy2020image} and employ decoders with identical architectures but independent parameters. Regarding data augmentation, we strictly follow the protocols established in UniMatch~\cite{yang2023revisiting}. Specifically, labeled images and ``weak'' unlabeled views undergo basic geometric transformations, including random resizing (scale range 0.5-2.0) and horizontal flipping. ``strong'' unlabeled views are generated via an aggressive perturbation chain comprising color jittering, random grayscale conversion, Gaussian blur, and CutMix~\cite{yun2019cutmix}. Furthermore, feature-space perturbation is applied via dropout~\cite{srivastava2014dropout} (rate $= 0.5$) within the dual-student forward mechanism. The unsupervised loss is guided by confidence-thresholded pseudo-labels (threshold $= 0.95$). Regarding the training schedule, each experiment runs for 80 epochs, with the exception of the MER dataset~\cite{li2022stepwise}. Due to its limited sample size, MER requires a prolonged schedule of 150 epochs to mitigate noisy updates and ensure stable convergence. All experiments were conducted on a workstation equipped with NVIDIA RTX 3090 GPUs. We utilize the mean intersection-over-union (mIoU) as the evaluation metric.
}

\subsection{Comparison with State-of-the-Art Methods}
\noindent
We report comparison results across six widely used remote-sensing benchmarks to evaluate the performance of the proposed Co2S comprehensively. We conduct extensive comparisons against the supervised baseline (``OnlySup"), which is trained exclusively on the labeled data, as well as several general semi-supervised baselines, including FixMatch~\cite{sohn2020fixmatch}, UniMatch~\cite{yang2023revisiting}, U2PL~\cite{wang2022semi}, and remote-sensing-specific approaches like WSCL~\cite{lu2023weak}, DWL~\cite{huang2024decouple}, and MUCA~\cite{wang2025semi}.

\begin{table}[t]
\centering
\small
\renewcommand{\arraystretch}{1.3} 

\caption{Segmentation results (mIoU \%) on the WHDLD dataset under different labeled ratios. The best and second-best results are highlighted in bold and underlined, respectively.}
\label{tab:whdld_results}
\newcommand{\pub}[1]{{\color{gray}~[#1]}}

\begin{tabular*}{\columnwidth}{@{\extracolsep{\fill}}ccccc}

\toprule 

Method & 1/24 & 1/16 & 1/8 & 1/4 \\

\midrule 

OnlySup & 53.6 & 55.2 & 58.0 & 60.5 \\
FixMatch~\cite{sohn2020fixmatch} \pub{NeurIPS'20} & 56.0 & 57.6 & 59.8 & 60.8 \\
U2PL~\cite{wang2022semi} \pub{CVPR'22} & 57.1 & 58.3 & 59.9 & 61.1 \\
WSCL~\cite{lu2023weak} \pub{TGRS'23} & 56.8 & 58.6 & 59.8 & 61.1 \\
UniMatch~\cite{yang2023revisiting} \pub{CVPR'23} & \underline{57.4} & \underline{58.8} & \underline{60.4} & \underline{61.5} \\
DWL~\cite{huang2024decouple} \pub{ISPRS'24} & 56.5 & 57.0 & 57.1 & 58.9 \\
MUCA~\cite{wang2025semi} \pub{TGRS'25} & 56.5 & 58.2 & 60.0 & 60.5 \\
\textbf{Co2S (Ours)} & \textbf{61.1} & \textbf{61.5} & \textbf{62.2} & \textbf{62.6} \\

\bottomrule 

\end{tabular*} 
\end{table}

\begin{table}[t]
\centering
\small
\renewcommand{\arraystretch}{1.3} 

\caption{Segmentation results (mIoU \%) on the LoveDA dataset under different labeled ratios. The best and second-best results are highlighted in bold and underlined, respectively.}
\label{tab:loveda_results}
\newcommand{\pub}[1]{{\color{gray}~[#1]}}

\begin{tabular*}{\columnwidth}{@{\extracolsep{\fill}}ccccc}
\toprule 
Method & 1/40 & 1/16 & 1/8 & 1/4 \\
\midrule 

OnlySup & 45.9 & 53.1 & 56.6 & 58.9 \\
FixMatch~\cite{sohn2020fixmatch} \pub{NeurIPS'20} & 53.3 & 57.9 & 60.1 & 61.5 \\
U2PL~\cite{wang2022semi} \pub{CVPR'22} & 52.3 & 58.0 & 59.6 & 61.5 \\
WSCL~\cite{lu2023weak} \pub{TGRS'23} & 54.1 & 57.4 & 59.4 & 61.1 \\
UniMatch~\cite{yang2023revisiting} \pub{CVPR'23} & 55.5 & 58.6 & 60.0 & 61.7 \\
DWL~\cite{huang2024decouple} \pub{ISPRS'24} & \underline{57.4} & \underline{59.0} & \textbf{62.8} & \underline{63.4} \\
MUCA~\cite{wang2025semi} \pub{TGRS'25} & 54.6 & 56.7 & 60.2 & 63.1 \\
\textbf{Co2S (Ours)} & \textbf{58.2} & \textbf{60.4} & \underline{62.7} & \textbf{64.0} \\

\bottomrule 
\end{tabular*} 
\end{table}

\begin{table}[t]
\centering
\small
\renewcommand{\arraystretch}{1.3}

\caption{Segmentation results (mIoU \%) on the Potsdam dataset under different labeled ratios. The best and second-best results are highlighted in bold and underlined, respectively.}
\label{tab:potsdam_results}

\newcommand{\pub}[1]{{\color{gray}~[#1]}}

\begin{tabular*}{\columnwidth}{@{\extracolsep{\fill}}ccccc}

\toprule 

Method & 1/32 & 1/16 & 1/8 & 1/4 \\

\midrule 

OnlySup & 61.5 & 64.9 & 70.2 & 73.8 \\
FixMatch~\cite{sohn2020fixmatch} \pub{NeurIPS'20} & 69.4 & 71.5 & 73.5 & 75.7 \\
U2PL~\cite{wang2022semi} \pub{CVPR'22} & 67.2 & 71.3 & 73.8 & 75.9 \\
WSCL~\cite{lu2023weak} \pub{TGRS'23} & 69.9 & 72.3 & 73.9 & 75.4 \\
UniMatch~\cite{yang2023revisiting} \pub{CVPR'23} & 70.7 & 72.6 & 74.8 & 76.4 \\
DWL~\cite{huang2024decouple} \pub{ISPRS'24} & \underline{74.2} & \textbf{78.3} & \textbf{79.8} & \textbf{80.3} \\
MUCA~\cite{wang2025semi} \pub{TGRS'25} & 66.9 & 71.5 & \underline{76.0} & 79.2 \\
\textbf{Co2S (Ours)} & \textbf{74.3} & \underline{76.6} & \textbf{79.8} & \underline{80.2} \\

\bottomrule 

\end{tabular*}
\end{table}
\begin{table}[t]
\centering
\small
\renewcommand{\arraystretch}{1.3}

\caption{Segmentation results (mIoU \%) on the MER and MSL datasets under different labeled ratios. The best and second-best results are highlighted in bold and underlined, respectively.}
\label{tab:mer_msl_miou}
\newcommand{\pub}[1]{{\color{gray}~[#1]}}

\begin{tabular*}{\columnwidth}{@{\extracolsep{\fill}}ccccc}
\toprule
\multirow{2}{*}{Method} & \multicolumn{2}{c}{MER} & \multicolumn{2}{c}{MSL} \\
\cmidrule(lr){2-3} \cmidrule(lr){4-5}
& 1/8 & 1/4 & 1/8 & 1/4 \\
\midrule
OnlySup & 52.8 & 55.3 & 53.6 & 59.0 \\
FixMatch~\cite{sohn2020fixmatch} \pub{NeurIPS'20} & 53.8 & \underline{58.1} & 56.8 & 60.5 \\
U2PL~\cite{wang2022semi} \pub{CVPR'22} & 54.9 & 56.1 & 55.5 & 61.6 \\
WSCL~\cite{lu2023weak} \pub{TGRS'23} & 51.9 & 54.9 & 58.2 & 59.9 \\
UniMatch~\cite{yang2023revisiting} \pub{CVPR'23} & \underline{56.0} & \underline{58.1} & 57.9 & 62.8 \\
DWL~\cite{huang2024decouple} \pub{ISPRS'24} & 54.5 & 54.7 & \underline{59.0} & \underline{63.6} \\
MUCA~\cite{wang2025semi} \pub{TGRS'25} & 52.6 & 53.0 & 52.4 & 59.3 \\
\textbf{Co2S (Ours)} & \textbf{56.8} & \textbf{59.1} & \textbf{60.9} & \textbf{65.9} \\
\bottomrule
\end{tabular*}
\end{table}











\begin{table*}[t] 
\centering 
\small 
\renewcommand{\arraystretch}{1.2} 

\caption{Segmentation results (mIoU \%) on the GID-15 dataset under different labeled ratios. The best and second-best results are highlighted in bold and underlined, respectively.}
\label{tab:gid_results}

\newcommand{\pub}[1]{{\color{gray}~[#1]}}

\begin{tabular*}{\textwidth}{ccccccccc} 
\toprule 

\multirow{2}{*}{Method} & 
\multirow{2}{*}{OnlySup} & 
FixMatch~\cite{sohn2020fixmatch} & 
U2PL~\cite{wang2022semi} & 
WSCL~\cite{lu2023weak} & 
UniMatch~\cite{yang2023revisiting} & 
DWL~\cite{huang2024decouple} & 
MUCA~\cite{wang2025semi} & 
\multirow{2}{*}{\textbf{Co2S (Ours)}} \\ 

 &  & \pub{NeurIPS'20} & \pub{CVPR'22} & \pub{TGRS'23} & \pub{CVPR'23} & \pub{ISPRS'24} & \pub{TGRS'25} & \\ 

\midrule 

1/8 & 67.5 & 70.0 & 67.2 & 72.8 & \underline{73.9} & 72.6 & 67.4 & \textbf{75.4} \\ 
1/4 & 74.1 & 74.8 & 75.3 & 75.4 & 75.9 & \underline{76.9} & 72.6 & \textbf{77.7} \\ 

\bottomrule 
\end{tabular*} 
\end{table*}

\subsubsection{Quantitative Comparison and Analysis}

\noindent
The comprehensive comparisons across six diverse benchmarks (Tables~\ref{tab:whdld_results}-\ref{tab:gid_results}) demonstrate that Co2S consistently achieves leading performance, particularly in regimes with extremely scarce annotations.  On the WHDLD~\cite{shao2020multilabel} dataset, Co2S outperforms the strong competitor UniMatch by 3.7\% under the extremely sparse 1/24 split, demonstrating its capability to parse diverse land-cover features accurately. Similarly, on the LoveDA~\cite{wang2021loveda} dataset, the method yields a remarkable 12.3\% improvement over the supervised baseline at the 1/40 ratio, showing superior performance against significant urban-rural distribution shifts. For the ultra-high-resolution Potsdam~\cite{isprs-annals-I-3-293-2012} dataset, Co2S achieves the best performance at the 1/32 split with 74.3\% mIoU, surpassing remote-sensing-specific designs like DWL. Extending to the large-scale GID-15~\cite{tong2020land} dataset, our method outperforms UniMatch by 1.5\% at the 1/8 split, indicating proficiency in handling diverse land-cover characteristics across vast spatial areas. Furthermore, in the extraterrestrial environments of MER and MSL~\cite{li2022stepwise}, Co2S delivers the highest accuracy on these benchmarks characterized by extreme class imbalance. Crucially, a consistent trend observed across all benchmarks is that the performance advantage of Co2S over baselines amplifies as the labeled ratio decreases. This phenomenon validates the stability of our Co2S, confirming that the synergistic foundation model priors effectively prevent model collapse and sustain high accuracy even in the most extreme low-data annotation regimes.

\subsubsection{Qualitative Visualization Analysis}
\begin{figure*}[t]
    \centering
    \includegraphics[width=\textwidth]{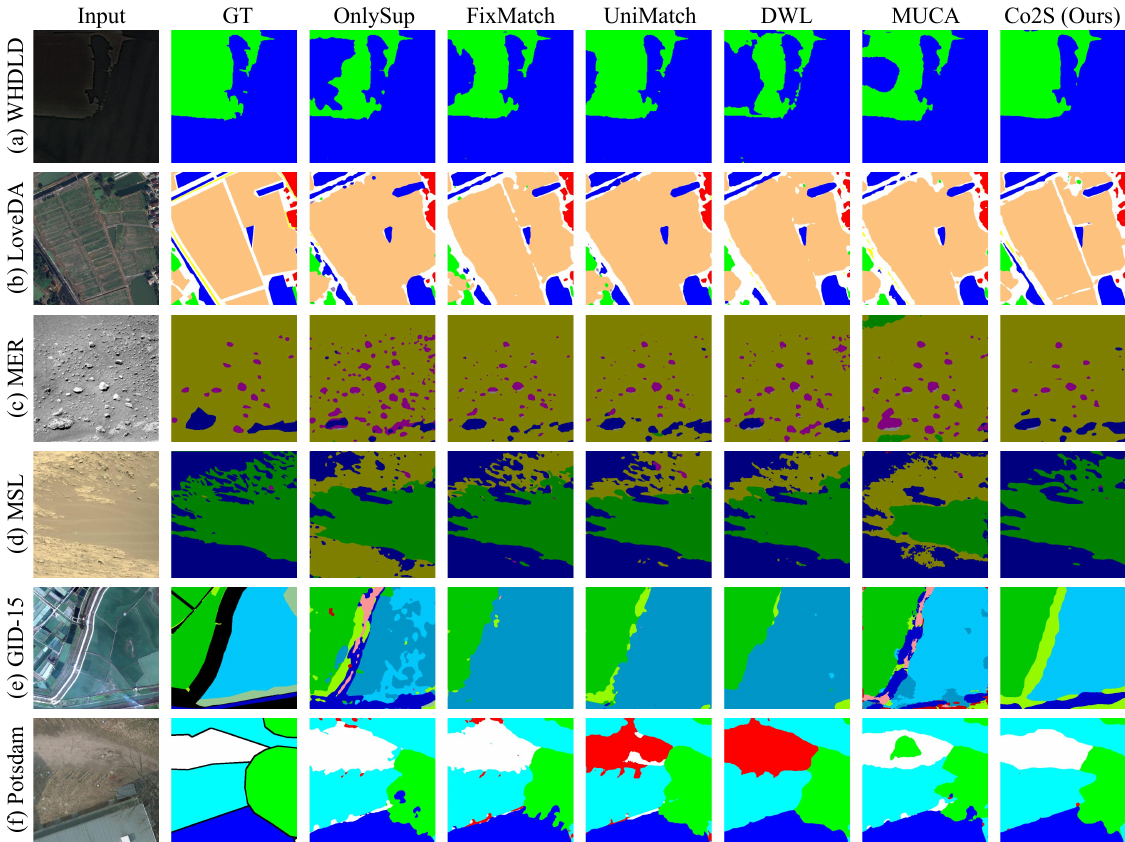}
    \caption{Visual comparison of semantic segmentation results on the six datasets under the 1/8 labeled ratio.}
    \label{fig:vis}
\end{figure*}

\noindent
To further substantiate these quantitative findings, we provide a visual comparison of segmentation results across all six datasets in Fig.~\ref{fig:vis}. As observed, baseline methods frequently struggle with semantic confusion in complex scenes. For instance, in the Potsdam dataset (Row f), UniMatch~\cite{yang2023revisiting} and DWL~\cite{huang2024decouple} incorrectly classify large areas of impervious surface (white) as clutter (red). Similarly, in WHDLD (Row a), the vegetation (green) in the top-left corner is misidentified as water (blue) by most competitors. Moreover, our method demonstrates superior capability in handling fine-grained objects and texture details. In the MER dataset (Row c), baselines suffer from severe over-segmentation, generating excessive false positives for scattered small rocks (purple), whereas Co2S precisely suppresses this noise to match the Ground Truth. Likewise, in MSL (Row d), other methods tend to misclassify large areas of bedrock (blue) as gravel (olive green), while Co2S accurately preserves the correct material distribution. These visualizations confirm that our framework effectively harmonizes global semantic consistency with local structural precision across diverse and challenging remote sensing environments, successfully rectifying category confusion while preserving fine-grained details.

\begin{figure}[t]
    \centering
    \includegraphics[width=\columnwidth]{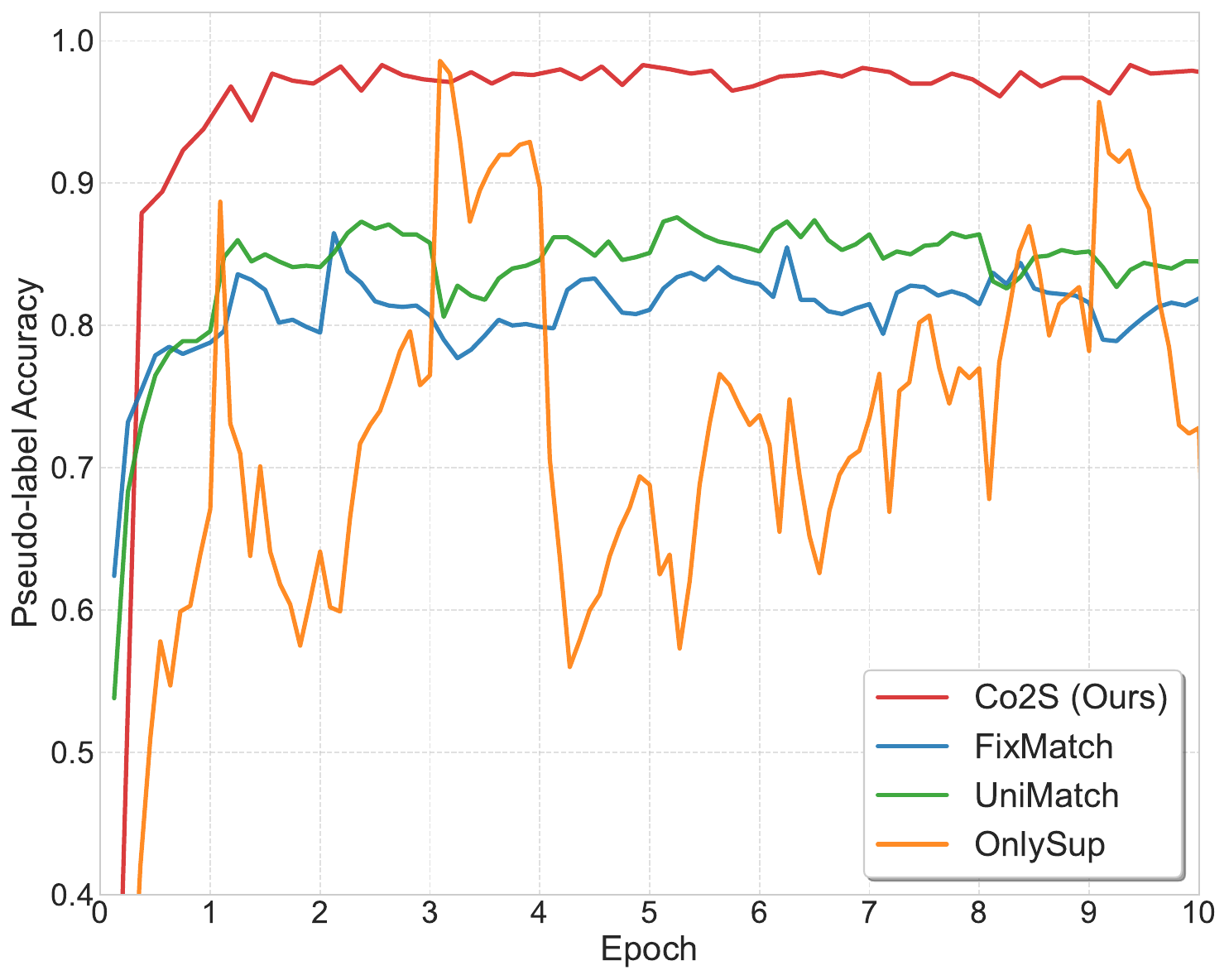}
    \caption{
         Evolution of pseudo-label accuracy during the early training phase (first 10 epochs) on the WHDLD dataset (1/24).  
    }
    \label{fig:pseudo_acc}
\end{figure}

\subsubsection{Pseudo-label Quality and Stability}

\noindent
To empirically verify our claim that Co2S effectively mitigates pseudo-label drift, we monitor the quality of pseudo-labels during the critical early training phase. The pseudo-label accuracy is calculated as the proportion of correctly classified pixels among those that pass the confidence threshold $\tau$:
\begin{equation}
\text{Accuracy} = \frac{\sum \mathbf{1}(\hat{\mathbf{y}}_u = \mathbf{y}_{gt}) \cdot \mathbf{1}(c \ge \tau)}{\sum \mathbf{1}(c \ge \tau)},
\end{equation}
where $\hat{\mathbf{y}}_u$ is the pseudo-label, $\mathbf{y}_{gt}$ is the ground truth, and $c$ is the confidence score. The evolution curves for the first 10 epochs on the WHDLD dataset (1/24 split) are plotted in Fig.~\ref{fig:pseudo_acc}. The results reveal distinct training dynamics:
1) The supervised baseline (OnlySup) exhibits extreme volatility, oscillating drastically between 50\% and 90\%. This highlights the instability of learning from scratch with scarce annotations.
2) Standard methods like FixMatch~\cite{sohn2020fixmatch} and UniMatch~\cite{sohn2020fixmatch} show signs of early saturation. They quickly plateau at a sub-optimal accuracy level ($\sim$80\%-88\%), suggesting that they are trapped by confirmation bias early in the training process and fail to correct initial errors.
3) In contrast, Co2S (red curve) demonstrates immediate semantic alignment. It surges to over 95\% accuracy within the very first epoch and maintains this superior level with minimal variance. 
This rapid convergence and sustained stability ($\sim$10\% gap over baselines) indicate that Co2S successfully establishes a high-quality supervision signal from the onset, thereby effectively stifling pseudo-label drift at its source and ensuring reliable optimization.

\subsection{Ablation Studies and Analysis}



\begin{table}[t]
\centering
\small
\renewcommand{\arraystretch}{1.2} 

\setlength{\tabcolsep}{25pt}

\caption{Ablation study on the explicit and implicit semantic guidance mechanisms (ESG: Explicit Semantic Guidance; ISG: Implicit Semantic Guidance) on the WHDLD dataset.}
\label{tab:co_guidance}

\begin{tabular}{ccc} 

\toprule
\begin{tabular}{@{}c@{}}ESG \end{tabular} &
\begin{tabular}{@{}c@{}}ISG \end{tabular} &
mIoU \\
\midrule
$\times$ & $\times$ & 58.97 \\
$\times$ & $\checkmark$ & 58.86 \\
$\checkmark$ & $\times$ & 60.77 \\
$\checkmark$ & $\checkmark$ & \textbf{61.09} \\
\bottomrule
\end{tabular}
\end{table}
\subsubsection{Impact of Explicit and Implicit Semantic Guidance}
\noindent{To validate the efficacy of the proposed explicit-implicit semantic co-guidance mechanism, we analyze the individual contributions of text embeddings and learnable queries which provide explicit and implicit class-level guidance respectively. As detailed in Table~\ref{tab:co_guidance}, the baseline model without guidance achieves 58.97 mIoU. Implementing implicit class-level guidance via learnable queries $\mathbf{Q}$ alone results in a slight performance drop to 58.86. This decline is attributed to the fact that the learnable queries are randomly initialized parameters lacking pre-trained priors. Optimizing these queries from scratch typically requires substantial annotated data, whereas the scarce labeled samples in our semi-supervised setting fail to provide sufficient supervision for their effective convergence. In contrast, integrating explicit class-level guidance via text embeddings $\mathbf{F}^t_{clip}$ alone yields a substantial gain of 1.8 and reaches 60.77 mIoU. This improvement stems from the inherent text-image alignment capability of CLIP acquired during pre-training, which enables the text embeddings to serve as powerful and reliable semantic anchors immediately. Notably, the optimal performance of 61.09 is attained only when both mechanisms are engaged. This result demonstrates a critical synergy where text embeddings ensure semantic correctness while learnable queries capture data-specific visual features, thereby jointly enhancing semantic consistency.}


\begin{table}[t]
\centering
\small
\renewcommand{\arraystretch}{1.2} 

\setlength{\tabcolsep}{16pt}

\caption{Ablation study on the heterogeneity of the dual-student architecture on the WHDLD dataset (1/24).}
\label{tab:co_fusion}

\begin{tabular}{cc} 
\toprule
Dual-student Architecture   & mIoU \\
\midrule
DINOv3 + DINOv3 (Local + Local)   & 45.20 \\
CLIP + CLIP (Global + Global )     & 60.78 \\
\textbf{CLIP + DINOv3 (Global + Local)}    & \textbf{61.09} \\
\bottomrule
\end{tabular}
\end{table}
\subsubsection{Benefit from Global and Local Feature Fusion}
\noindent{To verify the effectiveness of the proposed global-local feature collaborative fusion strategy, we investigate the necessity of the heterogeneous dual-student design by comparing it against homogeneous initialization strategies, as detailed in Table~\ref{tab:co_fusion}. First, the homogeneous configuration where both students utilize DINOv3 performs poorly, yielding an mIoU of only 45.20. This indicates that without strong global semantic priors, students initialized solely with local-structural features struggle to maintain category consistency, leading to severe semantic drift. Second, the homogeneous setting where both students employ CLIP attains a competitive mIoU of 60.78 due to rich global semantic representations. However, identical initialization results in highly correlated predictions. This lack of complementary viewpoints prevents the students from correcting each other's mistakes, rendering them prone to coupled error accumulation and confirmation bias. In contrast, our proposed heterogeneous architecture combining CLIP and DINOv3 achieves the highest performance of 61.09 mIoU. This superiority confirms that the framework successfully synergizes the global semantic representations of CLIP with the local structural details of DINOv3. By arbitrating between these complementary feature streams, Co2S ensures both semantic correctness and boundary precision, thereby stabilizing the semi-supervised training under scarce data conditions.}

\begin{table}[t]
\centering
\small
\renewcommand{\arraystretch}{1.2} 
\setlength{\tabcolsep}{20pt}

\caption{Ablation study on the contribution of different optimization objectives on the WHDLD dataset (1/24).}
\label{tab:loss_ablation}

\begin{tabular}{cccc} 
\toprule
$\mathcal{L}_{\text{sup}}$ & $\mathcal{L}_{\text{ct}}$ & $\mathcal{L}_{\text{sta}}$ & mIoU \\
\midrule
$\checkmark$ & $\times$ & $\times$ & 59.98 \\
$\checkmark$ & $\checkmark$ & $\times$ & 60.73 \\
$\checkmark$ & $\checkmark$ & $\checkmark$ & \textbf{61.09} \\
\bottomrule
\end{tabular}
\end{table}
\subsubsection{Effectiveness of Optimization Objectives}
\noindent
To evaluate the individual contribution of each term in our overall training objective, we conduct an ablation study as listed in Table~\ref{tab:loss_ablation}. The supervised baseline trained solely with supervised loss $\mathcal{L}_{\text{sup}}$ achieves 59.98 mIoU. Incorporating the consistency loss $\mathcal{L}_{\text{ct}}$ boosts the performance to 60.73, indicating that the weak-to-strong constraints effectively mine useful supervisory signals from unlabeled data to augment model training. Furthermore, adding the proposed stability loss $\mathcal{L}_{\text{sta}}$ yields the best performance of 61.09 mIoU. This further improvement confirms that $\mathcal{L}_{\text{sta}}$ is essential for mitigating pseudo-label drift by actively arbitrating between the two students based on confidence, thereby ensuring more reliable mutual learning.

\begin{table}[t]
\centering
\small 
\renewcommand{\arraystretch}{1.2} 

\setlength{\tabcolsep}{38pt}

\caption{Ablation study of synergizing CLIP with different SSM priors on the WHDLD dataset (1/24).}
\label{tab:whdld_encoder}
\begin{tabular}{cc} 
\toprule
SSM & mIoU  \\ 
\midrule
MAE~\cite{he2022masked}        & 60.84 \\
BEiTv2~\cite{peng2022beit}     & 60.98 \\
iBOT~\cite{zhou2021ibot}       & 60.96 \\
SimMIM~\cite{xie2022simmim}    & 60.99 \\
\textbf{DINOv3}~\cite{simeoni2025dinov3} & \textbf{61.09} \\
\bottomrule
\end{tabular}
\end{table}







\subsubsection{Synergizing CLIP with Different SSM Priors}
\noindent{To assess the advantage of heterogeneous model collaboration, we evaluate CLIP~\cite{radford2021learning} paired with a range of self-supervised models as well as with itself as a homogeneous baseline. The baseline attains an mIoU of 60.78, representing the performance ceiling achievable by relying solely on vision-language priors. As reported in Table~\ref{tab:whdld_encoder}, combining CLIP with any of the examined self-supervised models~\cite{he2022masked,peng2022beit,zhou2021ibot,xie2022simmim,simeoni2025dinov3} yields a performance gain over this baseline. The smallest improvement is 0.06 with MAE~\cite{he2022masked}, while the largest improvement is 0.31 with DINOv3~\cite{simeoni2025dinov3}. This consistent uplift across all pairings confirms that complementary visual priors from diverse self-supervised paradigms substantially enhance mutually semantic interactive learning beyond what CLIP alone can deliver.}

\section{Conclusion}
\noindent
We propose Co2S, a stable semi-supervised framework for remote sensing segmentation that leverages complementary priors from two heterogeneous vision foundation models: a vision-language model (e.g., CLIP) and a self-supervised model (e.g., DINOv3). By co-guiding and co-fusing their inherently distinct representations, Co2S establishes a drift-resistant mutual learning process that simultaneously ensures semantic consistency and boundary accuracy. Experiments on six benchmarks demonstrate state-of-the-art performance, with especially pronounced gains under extreme label scarcity.

{
\bibliographystyle{IEEEtran}
\bibliography{refs}
}

\begin{IEEEbiography}[{\includegraphics[width=1in,height=1.25in, clip,keepaspectratio]{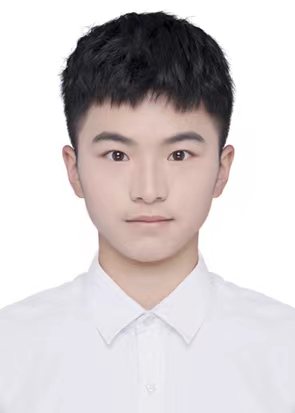}}]{Yi Zhou} is pursuing the B.E. degree with the School of Computer Technology and Application at Qinghai University, Xining, China. His research interests primarily focus on computer vision, particularly in the field of remote sensing image processing.
\end{IEEEbiography}
\vspace{-8mm}

\begin{IEEEbiography}[{\includegraphics[width=1in,height=1.25in, clip,keepaspectratio]{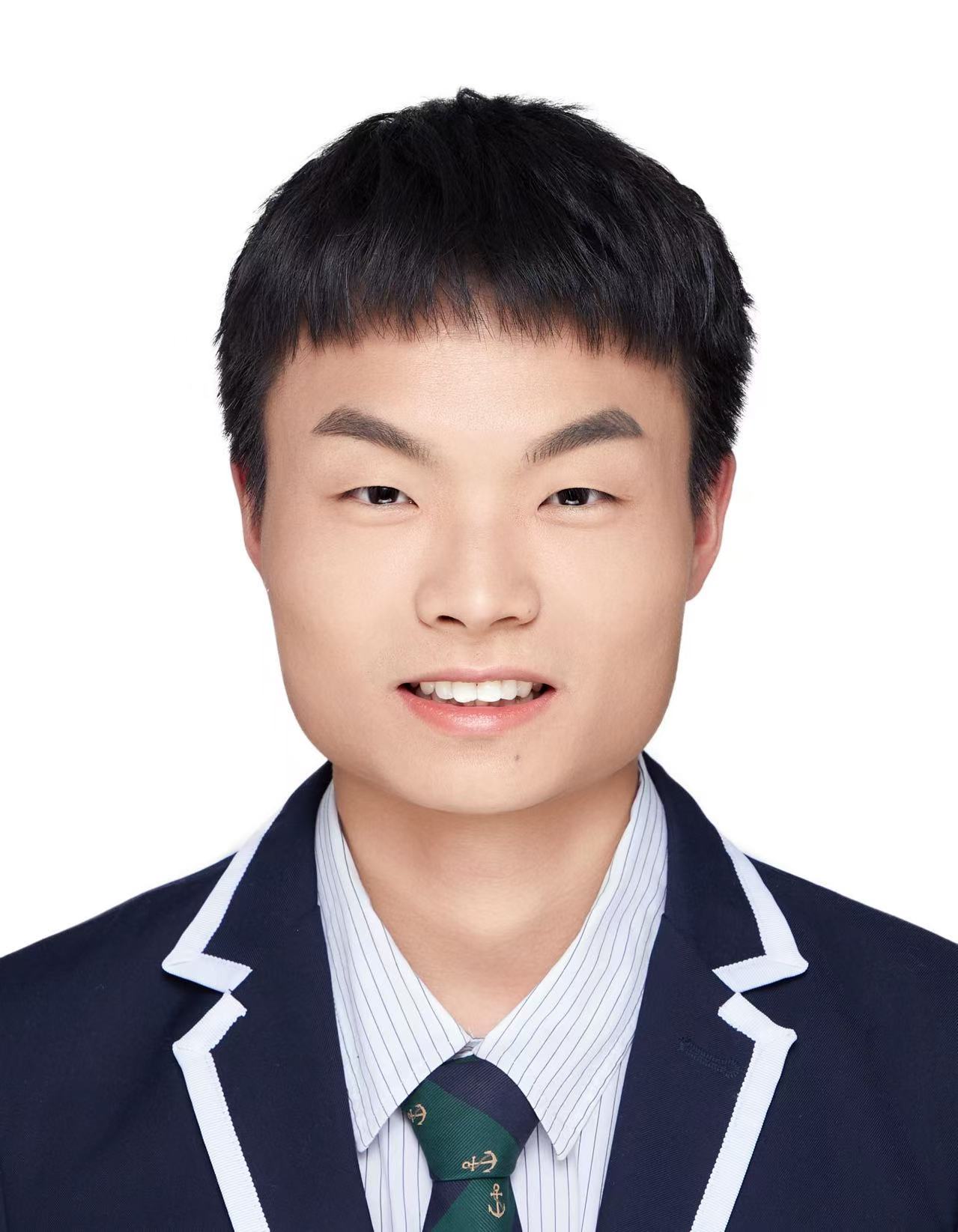}}]{Xuechao Zou} received the B.E. degree in 2021 and the M.S. degree in 2024 from the School of Computer Technology and Application at Qinghai University, Xining, China. He is currently pursuing a Ph.D. degree with the School of Computer Science and Technology at Beijing Jiaotong University. His research interests focus on computer vision, particularly remote sensing image processing.
\end{IEEEbiography}
\vspace{-8mm}

\begin{IEEEbiography}[{\includegraphics[width=1in,height=1.25in, clip,keepaspectratio]{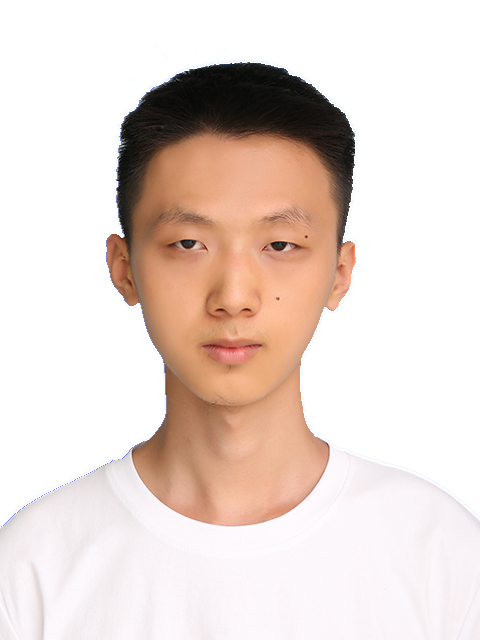}}]{Shun Zhang} is pursuing the B.E. degree with the School of Computer Technology and Application at Qinghai University, Xining, China. He is currently pursuing the M.S. degree with the School of Computer Science
and Technology at Beijing Jiaotong University. His research interests primarily focus on computer vision, particularly in the field of remote sensing image processing.
\end{IEEEbiography}
\vspace{-8mm}

\begin{IEEEbiography}[{\includegraphics[width=1in,height=1.25in, clip,keepaspectratio]{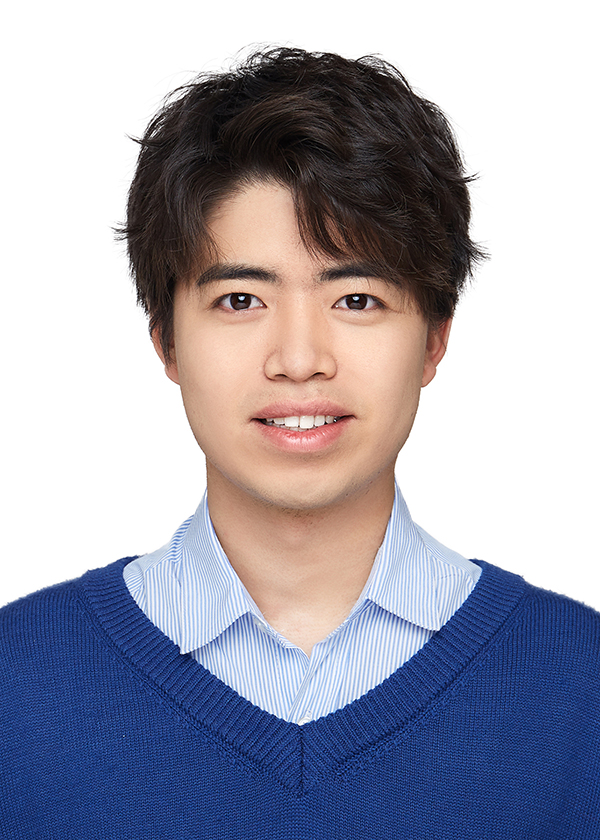}}]{Kai Li} received his B.E. degree from the School of Computer Technology and Application at Qinghai University, Xining, China, in 2020, and his M.S. degree from the Department of Computer Science and Technology at Tsinghua University, Beijing, China, in 2024. He is pursuing a Ph.D. in the Department of Computer Science and Technology at Tsinghua University. His research interests focus primarily on speech separation and audio-visual learning.
\end{IEEEbiography}
\vspace{-8mm}

\begin{IEEEbiography}[{\includegraphics[width=1in,height=1.25in, clip,keepaspectratio]{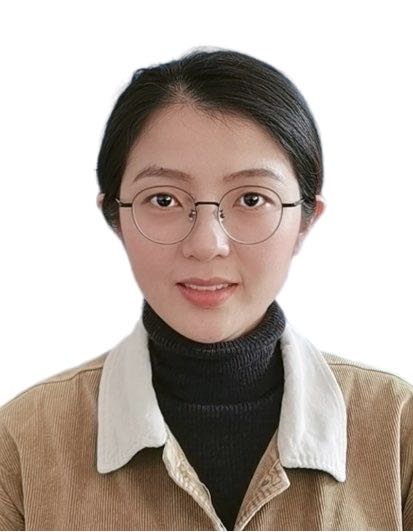}}]{Shiying Wang} received the B.E. degree from Jilin Agricultural University, Changchun, China, in 2015, the M.S. degree and Ph.D. degree from the School of Computer Technology and Application, Qinghai University, Xining, China, in 2018 and 2025. Her research interests mainly include remote sensing image processing, particularly in grassland informatics.
\end{IEEEbiography}
\vspace{-8mm}

\begin{IEEEbiography}[{\includegraphics[width=1in,height=1.25in, clip,keepaspectratio]{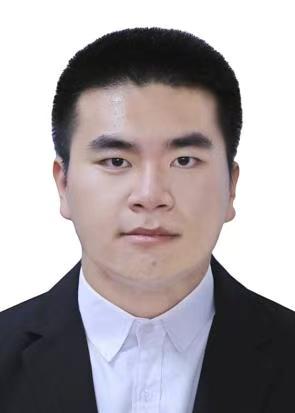}}]{Jingming Chen} is pursuing the B.E. degree with the School of Computer Technology and Application at Qinghai University, Xining, China. His research interests primarily focus on computer vision, particularly in the field of remote sensing image processing.
\end{IEEEbiography}
\vspace{-8mm}

\begin{IEEEbiography}[{\includegraphics[width=1in,height=1.25in, clip,keepaspectratio]{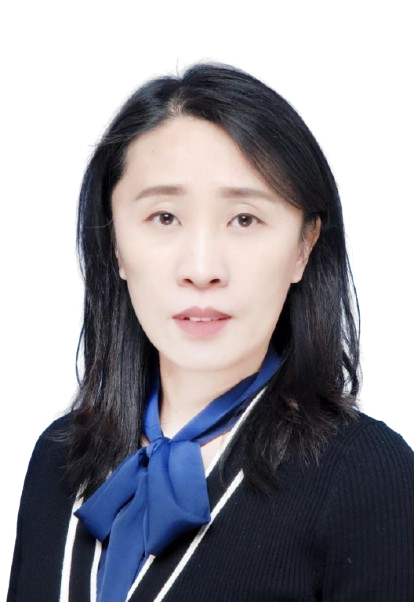}}]{Congyan Lang} received the Ph.D. degree from the Beijing Key Laboratory of Traffic Data Analysis and Mining, School of Computer and Information Technology, Beijing Jiaotong University, Beijing, China, in 2006. She was a Visiting Professor with the Department of Electrical and Computer Engineering, National University of Singapore, Singapore, from 2010 to 2011. From 2014 to 2015, she was a visiting professor at the Department of Computer Science, University of Rochester, Rochester, NY, USA. She is a Professor at the School of Computer and Information Technology, Beijing Jiaotong University. She has published over 80 research articles in various journals and refereed conferences. Her research areas include computer vision and machine learning.
\end{IEEEbiography}
\vspace{-8mm}

\begin{IEEEbiography}[{\includegraphics[width=1in,height=1.25in, clip,keepaspectratio]{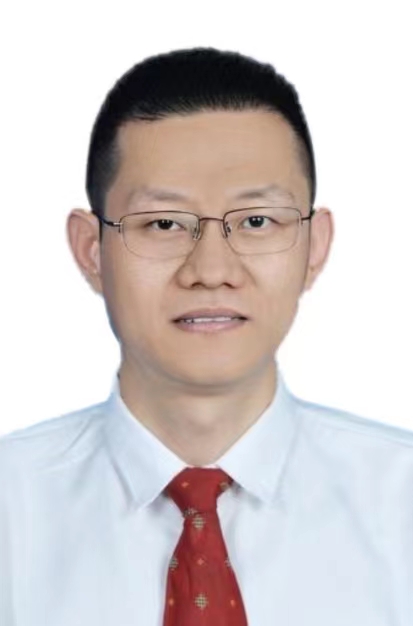}}]{Tengfei Cao}  received the Ph.D. degree from Beijing University of Posts and Telecommunications (BUPT), Beijing, China, in July 2020. He is currently an Associate Professor with the Department of Computer Technology and Applications, Qinghai University, Xining, China. He has published several high-quality research papers, such as ACM Transactions on Multimedia Computing, Communications, and Applications, IEEE TRANSACTIONS ON NETWORK AND SERVICE
MANAGEMENT, and CHINA COMMUNICATIONS. His research interests include multimedia communications and privacy security.
\end{IEEEbiography}
\vspace{-8mm}

\begin{IEEEbiography}[{\includegraphics[width=1in,height=1.25in, clip,keepaspectratio]{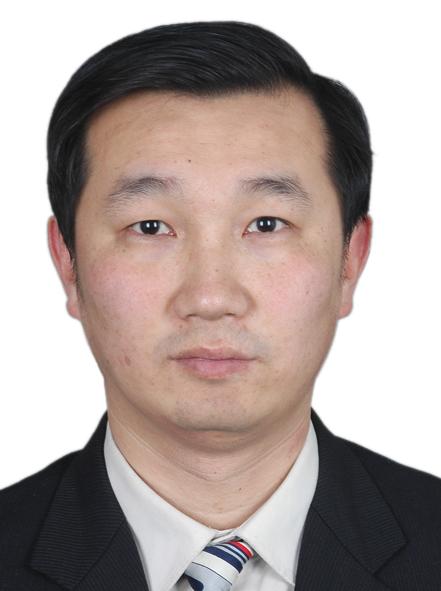}}]{Pin Tao} received his B.S. degree and Ph.D. in computer science and technology from Tsinghua University, Beijing, China, in 1997 and 2002. He is an Associate Professor at the Computer Science and Technology Department, Tsinghua University, Beijing, China. He is also the vice director of the Key Laboratory of Pervasive Computing,  Ministry of Education. Dr. Tao has published more than 80 papers and over 10 patents. His current research interests mainly focus on human-AI hybrid intelligence and multimedia-embedded processing.
\end{IEEEbiography}

\begin{IEEEbiography}[{\includegraphics[width=1in,height=1.25in, clip,keepaspectratio]{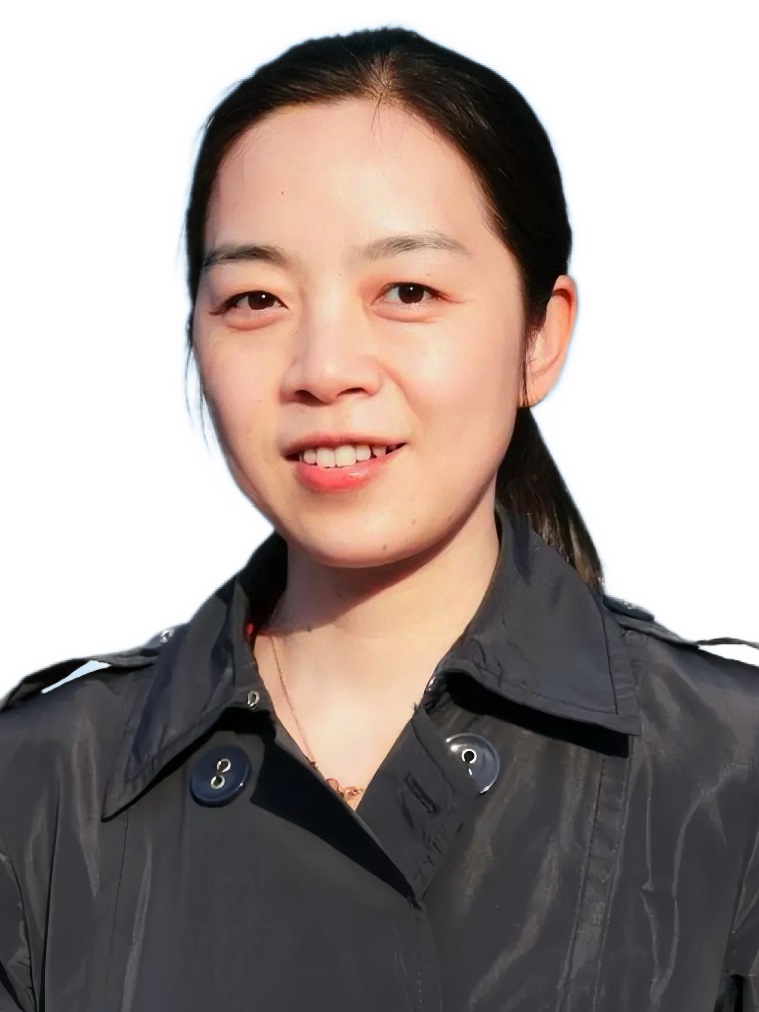}}]{Yuanchun Shi} 
received the PhD, MS and BS degrees in computer science from Tsinghua University. She is a Changjiang distinguished professor at the Department of Computer Science, Tsinghua University. She was a senior visiting scholar at the MIT AI Lab during 2001-2002. She had chaired several conferences, including ACM Ubicomp2011. She serves as the area editor of pervasive and mobile computing (PMC, Elsevier), editor of interacting with computer (IwC, Oxford University Press), and vice editor-in-chief of Communications of China Computer Federation (CCF). She has published more than one hundred papers in IJHCS, IEEE TPDS, TKDE, ACM CHI, MM, UIST, etc. Her research interests include human-computer interaction, pervasive computing, and multimedia communication. She is a Fellow of the IEEE.
\end{IEEEbiography}
 
\vfill

\end{document}